\documentclass[10pt,twocolumn,letterpaper]{article}

\usepackage{iccv}
\usepackage{times}
\usepackage{epsfig}
\usepackage{graphicx}
\usepackage{amsmath}
\usepackage{amssymb}
\usepackage{booktabs}
\usepackage[symbol]{footmisc}

\usepackage[pagebackref=true,breaklinks=true,letterpaper=true,colorlinks,bookmarks=false]{hyperref}

\iccvfinalcopy 

\def\iccvPaperID{2548} 

\ificcvfinal\fi

\begin{document}


\title{SnowflakeNet: Point Cloud Completion by Snowflake Point Deconvolution with Skip-Transformer}

\author{Peng Xiang\textsuperscript{1}\footnotemark[1], Xin Wen\textsuperscript{1,4}\footnotemark[1], Yu-Shen Liu\textsuperscript{1}, Yan-Pei Cao\textsuperscript{2}, Pengfei Wan\textsuperscript{2}, Wen Zheng\textsuperscript{2}, Zhizhong Han\textsuperscript{3}\\
\textsuperscript{1}School of Software, BNRist, Tsinghua University, Beijing, China\\
\textsuperscript{2}Y-tech, Kuaishou Technology, Beijing, China\hspace{3mm}
\textsuperscript{3}Wayne State University\hspace{3mm}\textsuperscript{4}JD.com, Beijing, China\\
{\small xp20@mails.tsinghua.edu.cn\hspace{3mm}wenxin16@jd.com\hspace{3mm}liuyushen@tsinghua.edu.cn\hspace{1mm}}\\
{\small caoyanpei@gmail.com\hspace{3mm}\{wanpengfei,zhengwen\}@kuaishou.com\hspace{3mm}h312h@wayne.edu}

}

\begin{figure}[t]

\twocolumn[{%
\maketitle
\begin{center}
    \centering
    \vspace{-7mm}
    \includegraphics[width=1\textwidth]{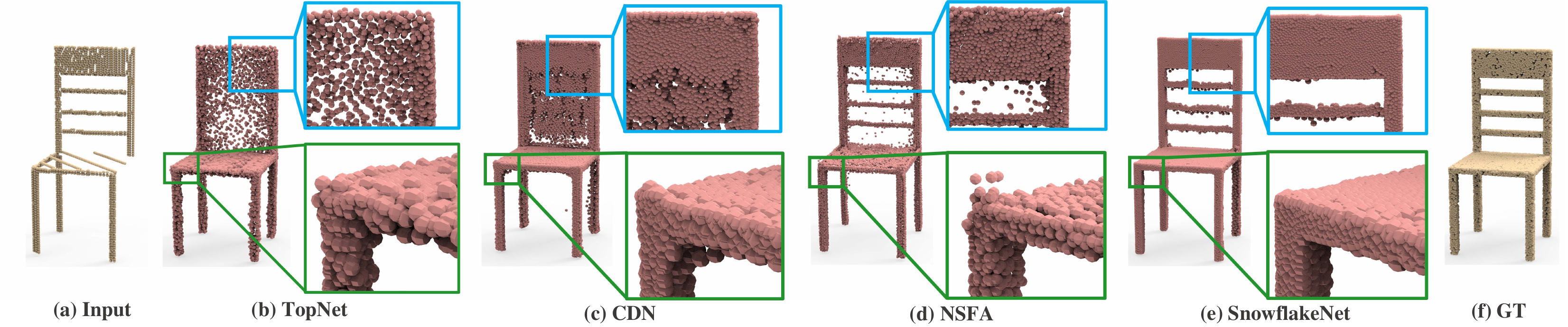}
    \caption{Visual comparison of point cloud completion results. The input and ground truth have 2048 and 16384 points, respectively. Compared with current completion methods like TopNet \cite{tchapmi2019topnet}, CDN \cite{wang2020cascaded} and NSFA \cite{zhang2020detail}, our SnowflakeNet can generate the complete shape (16384 points) with fine-grained geometric details, such as smooth regions (blue boxes), sharp edges and corners (green boxes).}
    \label{fig:teaser}
\end{center}%
}]
\vspace{-6mm}
\end{figure}

\maketitle

\renewcommand{\thefootnote}{\fnsymbol{footnote}}
\footnotetext[1]{Equal contribution. This work was supported by National Key R\&D Program of China (2020YFF0304100), the National Natural Science Foundation of China (62072268), and in part by Tsinghua-Kuaishou Institute of Future Media Data. The corresponding author is Yu-Shen Liu.} 

\ificcvfinal\thispagestyle{empty}\fi


\begin{abstract}
\vspace{-2mm}
Point cloud completion aims to predict a complete shape in high accuracy from its partial observation. However, previous methods usually suffered from discrete nature of point cloud and unstructured prediction of points in local regions, which makes it hard to reveal fine local geometric details on the complete shape. To resolve this issue, we propose SnowflakeNet with Snowflake Point Deconvolution (SPD) to generate the complete point clouds. The SnowflakeNet models the generation of complete point clouds as the snowflake-like growth of points in 3D space, where the child points are progressively generated by splitting their parent points after each SPD. Our insight of revealing detailed geometry is to introduce skip-transformer in SPD to learn point splitting patterns which can fit local regions the best. Skip-transformer leverages attention mechanism to summarize the splitting patterns used in the previous SPD layer to produce the splitting in the current SPD layer. The locally compact and structured point cloud generated by SPD is able to precisely capture the structure characteristic of 3D shape in local patches, which enables the network to predict highly detailed geometries, such as smooth regions, sharp edges and corners. Our experimental results outperform the state-of-the-art point cloud completion methods under widely used benchmarks. Code will be available at \url{https://github.com/AllenXiangX/SnowflakeNet}.

\end{abstract}

\vspace{-6mm}
\section{Introduction}
In 3D computer vision \cite{han2020shapecaptioner, han2019view, han20182seq2seq, han2018deep, han2017boscc} applications, raw point clouds captured by 3D scanners and depth cameras are usually sparse and incomplete \cite{wen2020sa,wen2020pmp,wen2020cycle} due to occlusion and limited sensor resolution. Therefore, 
point cloud completion \cite{wen2020sa,tchapmi2019topnet}, which aims to predict a complete shape from its partial observation, is vital for various downstream tasks.
%
Benefiting from large-scaled point cloud datasets, deep learning based point cloud completion methods have been attracting more research interests. Current methods either constrain the generation of point clouds by following a hierarchical rooted tree structure \cite{wang2020cascaded,xie2020grnet,tchapmi2019topnet} or assume a specific topology \cite{yang2018foldingnet,wen2020sa} for the target shape. However, most of these methods suffered from discrete nature of point cloud and unstructured prediction of points in local regions, which makes it hard to preserve a well arranged structure for points in local patches. It is still challenging to capture the local geometric details and structure characteristic on the complete shape, such as smooth regions, sharp edges and corners, as illustrated in Figure \ref{fig:teaser}. 

\begin{figure}[t]
\begin{center}
   \includegraphics[width=\linewidth]{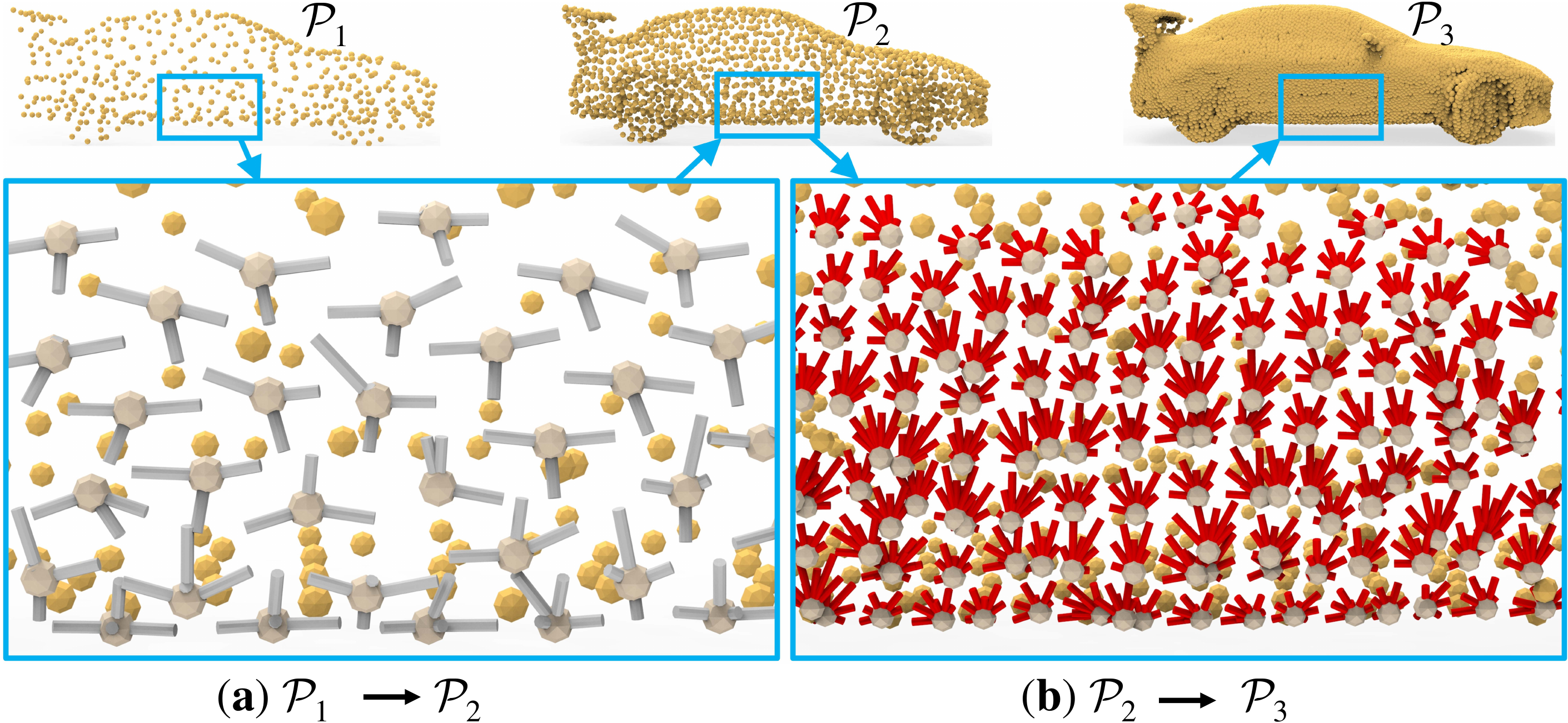}
   \caption{Illustration of snowflake point deconvolution (SPD) for growing part of a car. To show the local change more clearly, we only illustrate some sample points as parent points in the same patch and demonstrate their splitting paths for child points, which are marked as gray and red lines. (a) illustrates the SPD of point splitting from a coarse point cloud $\mathcal{P}_1$ (512 points) to its splitting $\mathcal{P}_2$ (2048 points). (b) illustrates the SPD of point splitting from $\mathcal{P}_2$ to dense complete point cloud $\mathcal{P}_3$ (16384 points), where the child points are expanding like the growth process of snowflakes.}
\label{fig:snowflake_conv}
\end{center}
\vspace{-8mm}
\end{figure}

In order to address this problem, we propose a novel network called \emph{SnowflakeNet}, especially focusing on the decoding process to complete partial point clouds. SnowflakeNet mainly consists of layers of \emph{Snowflake Point Deconvolution} (SPD), which models the generation of complete point clouds like the snowflake growth of points in 3D space. We progressively generate points by stacking one SPD layer upon another, where each SPD layer produces child points by splitting their parent point with inheriting shape characteristics captured by the parent point. Figure \ref{fig:snowflake_conv} illustrates the process of SPD and point-wise splitting. 

Our insight of revealing  detailed  geometry is to introduce \emph{skip-transformer} in SPD to learn point splitting patterns which can fit local regions the best. Compared with the previous methods, which often ignore the spatial relationship among points \cite{yang2018foldingnet,tchapmi2019topnet,liu2020morphing} or simply learn through self-attention in a single level of multi-step point cloud decoding \cite{wen2020sa,li2018pu,wang2020cascaded}, our skip-transformer is proposed to integrate the spatial relationships across different levels of decoding. Therefore, it can establish a cross-level spatial relationships between points in different decoding steps, and refine their location to produce more detailed structure. To achieve this, skip-transformer leverages attention mechanism to summarize the splitting patterns used in the previous SPD layer, which aims to produce the splitting in current SPD layer. The skip-transformer can learn the shape context and the spatial relationship between the points in local patches. This enables the network to precisely capture the structure characteristic in each local patches, and predict a better point cloud shape for both smooth plane and sharp edges in 3D space. We achieved the state-of-the-art completion accuracy under the widely used benchmarks. 
Our main contributions can be summarized as follows.

\vspace{-1mm}
\begin{itemize}
  \vspace{-2mm}
  \item We propose a novel SnowflakeNet for point cloud completion. Compared with previous locally unorganized complete shape generation methods, SnowflakeNet can interpret the generation process of complete point cloud into an explicit and locally structured pattern, which greatly improves the performance of 3D shape completion.
  \vspace{-2mm}
  \item We propose the novel Snowflake Point Deconvolution (SPD) for progressively increasing the number of points. It reformulates the generation of child points from parent points as a growing process of snowflake, where the shape characteristic embedded by the parent point features is extracted and inherited into the child points through a \emph{point-wise splitting} operation.
  \vspace{-2mm}
  \item We introduce a novel skip-transformer to learn splitting patterns in SPD. It learns shape context and spatial relationship between child points and parent points, which encourages SPD to produce locally structured and compact point arrangements, and capture the structure characteristic of 3D surface in local patches.
\end{itemize}
\vspace{-2mm}
%
%
\begin{figure*}
\begin{center}
\includegraphics[width=\textwidth]{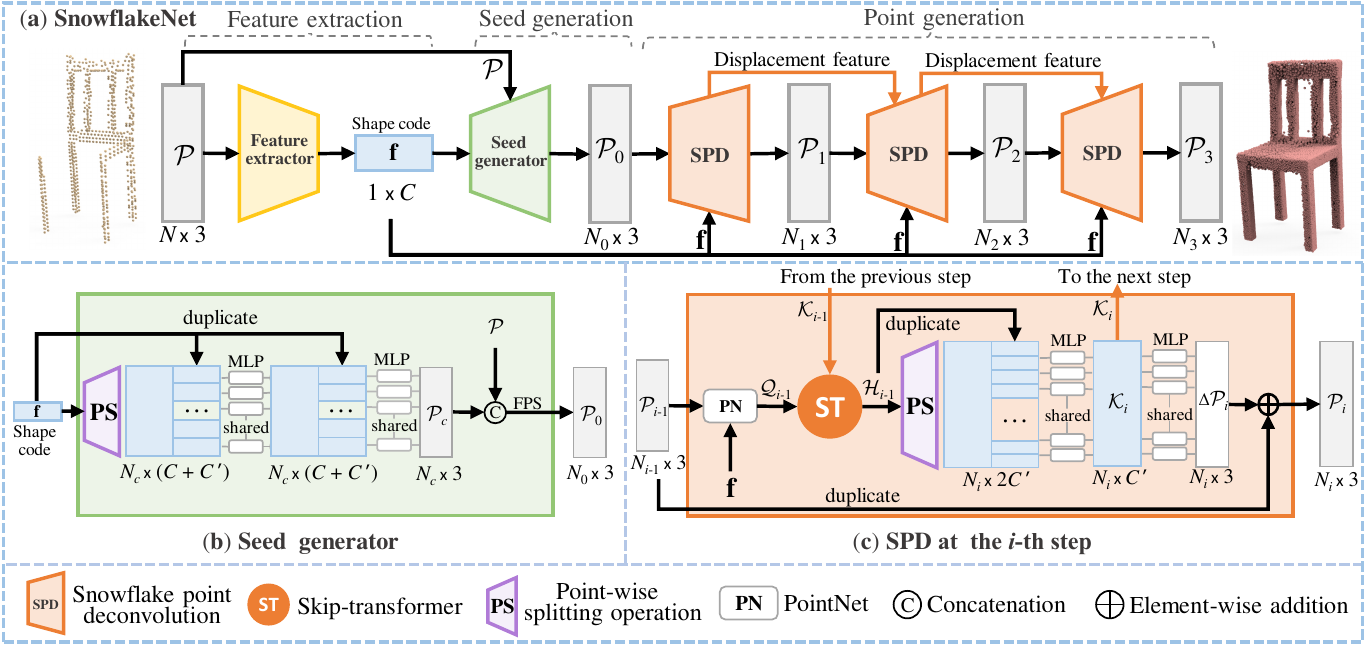}
\end{center}
    \vspace{-2mm}
   \caption{ (a) The overall architecture of SnowflakeNet, which consists of three modules: feature extraction, seed generation and point generation. 
   (b) The details of seed generation module. (c) Snowflake point deconvolution (SPD). Note that $N$, $N_c$ and $N_i$ are the number of points, $C$ and $C'$ are the number of point feature channels that are 512 and 128, respectively.}
\label{fig:overall}
\vspace{-5mm}
\end{figure*}
\vspace{-3mm}
\section{Related Work}
\vspace{-1mm}
Point cloud completion methods can be roughly divided into two categories. (1) Traditional point cloud completion methods \cite{sung2015data,berger2014state,thanh2016field,wei2019local} usually assume a smooth surface of 3D shape, or utilize large-scaled complete shape dataset to infer the missing regions for incomplete shape. (2) Deep learning \cite{Jiang2019SDFDiffDRcvpr, han2019parts4feature, han20193d2seqviews, han20193dviewgraph, han2018seqviews2seqlabels, han2020seqxy2seqz, han2020reconstructing, wen2020hvpredictor, wen2020cmpd, wen2019adversarial} based methods \cite{huang2020pf,wang2020cascaded,chen2019unpaired, gu2020weakly, NEURIPS2020_ba036d22, hutaoaaai2020, hu2019render4completion}, however, learn to predict a complete shape based on the learned prior from the training data. Our method falls into the second class and focuses on the decoding process of point cloud completion. We briefly review deep learning based methods below.

\noindent\textbf{Point cloud completion by folding-based decoding.} The development of deep learning based 3D point cloud processing techniques \cite{han2020drwr, wen2020cf, wen2020point2spatialcapsule, liu2019l2g, liu2019point2sequence, liu2021fine, liu2020lrc, han2019multi, cc2021matching, NeuralPull} have boosted the research of point cloud completion.
Suffering from the discrete nature of point cloud data, the generation of high-quality complete shape is one of the major concerns in the point cloud completion research.
One of the pioneering work is the FoldingNet \cite{yang2018foldingnet}, although it was not originally designed for point cloud completion. It proposed a two-stage generation process and combined with the assumption that 3D object lies on 2D-manifold \cite{tchapmi2019topnet}. Following the similar practice, methods like SA-Net \cite{wen2020sa} further extended such generation process into multiple stages by proposing hierarchical folding in decoder. However, the problem of these folding-based methods \cite{wen2020sa,yang2018foldingnet,li2019pu} is that the 3-dimensional code generated by intermediate layer of network is an implicit representation of target shape, which 
is hardly interpreted or constrained in order to help refine the shape in local region.
On the other hand, TopNet \cite{tchapmi2019topnet} modeled the point cloud generation process as the growth of rooted tree, where one parent point feature is projected into several child point features in a feature expansion layer of TopNet. Same as FoldingNet \cite{yang2018foldingnet}, the intermediate generation processes of TopNet and SA-Net are also implicit, where the shape information is only represented by the point features, cannot be constrained or explained explicitly.

\noindent\textbf{Point cloud completion by coarse-to-fine decoding.} Recently, explicit coarse-to-fine completion framework \cite{xie2020grnet,dai2017shape} has received an increasing attention, due to its explainable nature and controllable generation process. Typical methods like PCN \cite{yuan2018pcn} and NSFA \cite{zhang2020detail} adopted the two-stage generation framework, where a coarse and low resolution point cloud is first generated by the decoder, and then a lifting module is used to increase the density of point clouds. Such kind of methods can achieve better performance since it can impose more constraints on the generation process of point cloud, i.e. the coarse one and the dense one. Followers like CDN \cite{wang2020cascaded} and PF-Net \cite{huang2020pf} further extended the number of generation stages and achieved the currently state-of-the-art performance. Although intriguing performance has been achieved by the studies along this line, most of these methods still cannot predict a locally structured point splitting pattern, as illustrated in Figure \ref{fig:teaser}. The biggest problem is that these methods only focus on the expansion of point number and the reconstruction of global shape, while ignoring to preserve a well-structured generation process for points in local regions. This makes these methods difficult to capture local detailed geometries and structures of 3D shape.

Compared with the above-mentioned methods, our SnowflakeNet takes one step further to explore an explicit, explainable and locally structured solution for the generation of complete point cloud. SnowflakeNet models the progressive generation of point cloud as a hierarchical rooted tree structure like TopNet, while keeping the process explainable and explicit like CDN \cite{wang2020cascaded} and PF-Net \cite{huang2020pf}. Moreover, it excels the predecessors by arranging the point splitting in local regions in a locally structured pattern, which enables to precisely capture the detailed geometries and structures of 3D shapes.

\noindent\textbf{Relation to transformer.} Transformer \cite{vaswani2017attention} was initially proposed for encoding sentence in natural language processing, and soon gets popular in the research of 2D computer vision (CV) \cite{dosovitskiy2021an,parmar2018image}. Then, the success of transformer-based 2D CV studies have drawn the attention of 3D point cloud research, where pioneering studies like Point-Transformer \cite{zhao2021pointtransformer}, PCT \cite{Guo_2021} and Pointformer \cite{Pan_2021_CVPR} have introduced such framework in the encoding process of point cloud to learn the representation. In our work, instead of only utilizing its representation learning ability, we further extend the application of transformer-based structure into the decoding process of point cloud completion, and reveal its ability for generating high quality 3D shapes through the proposed skip-transformer.
\vspace{-2mm}
\section{SnowflakeNet}
The overall architecture of SnowflakeNet is shown in Figure \ref{fig:overall}(a), which consists of three modules: feature extraction, seed generation and point generation. We will detail each module in the following. 

%
\subsection{Overview}
\noindent\textbf{Feature extraction module.}
Let $\mathcal{P} = \{ \mathbf{p}_j \}$ of size $N \times 3$ be an input point cloud, where $N$ is the number of points and each point $\mathbf{p}_j$ indicates a 3D coordinate. The feature extractor aims to extract a shape code $\mathbf{f}$ of size $1 \times C$, which captures the global structure and detailed local pattern of the target shape. To achieve this, we adopt three layers of set abstraction from \cite{qi2017pointnet++} to aggregate point features from local to global, along which point transformer \cite{zhao2021pointtransformer} is applied to incorporate local shape context.


\noindent\textbf{Seed generation module.}
The objective of the seed generator is to produce a coarse but complete point cloud $\mathcal{P}_0$ of size $N_0 \times 3$ that captures the geometry and structure of the target shape. As shown in Figure \ref{fig:overall}(b), with the extracted shape code $\mathbf{f}$, the seed generator first produces point features that capture both the existing and missing shape through point-wise splitting operation. Next, the per-point features are integrated with the shape code through multi-layer perceptron (MLP) to generate a coarse point cloud $\mathcal{P}_c$ of size $N_c \times 3$. Then, following the previous method \cite{wang2020cascaded}, $\mathcal{P}_c$ is merged with the input point cloud $\mathcal{P}$ by concatenation, and then the merged point cloud is down-sampled to $\mathcal{P}_0$ through farthest point sampling (FPS) \cite{qi2017pointnet++}.
In this paper, we typically set $N_c = 256$ and $N_0=512$, where a sparse point cloud $\mathcal{P}_0$ suffices for representing the underlying shape. $\mathcal{P}_0$ will serve as the seed point cloud for point generation module.

\noindent\textbf{Point generation module.}
The point generation module consists of three steps of Snowflake Point Deconvolution (SPD), each of which takes the point cloud from the previous step and splits it by up-sampling factors (denoted by $r_1, r_2 $ and $r_3$) to obtain $\mathcal{P}_1, \mathcal{P}_2$ and $ \mathcal{P}_3$, which have the point sizes of $N_1 \times 3, N_2 \times 3$ and $N_3 \times 3$, respectively. SPDs collaborate with each other to generate a rooted tree structure that complies with local pattern for every seed point. The structure of SPD is detailed below.


\subsection{Snowflake Point Deconvolution (SPD)}

The SPD aims to increase the number of points by splitting each parent point into multiple child points, which can be achieved by first duplicating the parent points and then adding variations. Existing methods \cite{wang2020cascaded,yuan2018pcn,zhang2020detail} usually adopt the folding-based strategy \cite{yang2018foldingnet} to obtain the variations, which are used for learning different displacements for the duplicated points. However, the folding operation samples the same 2D grids for each parent point, which ignores the local shape characteristics contained in the parent point.
Different from the folding-based methods \cite{wang2020cascaded,yuan2018pcn,zhang2020detail}, the SPD obtains variations through a \textit{point-wise splitting} operation, which fully leverages the geometric information in parent points and adds variations that comply with local patterns. 
In order to progressively generate the split points, three SPDs are used in point generation module. In addition, to facilitate consecutive SPDs to split points in a coherent manner, we propose a novel skip-transformer to 
capture the shape context and the spatial relationship between the parent points and their split points.

Figure \ref{fig:overall}(c) illustrates the structure of the $i$-th SPD with up-sampling factor $r_i$. We denote a set of parent points obtained from previous step as $\mathcal{P}_{i-1}=\{ \mathbf{p}_j^{i-1}\}_{j=1}^{N_{i-1}}$.
We split the parent points in $\mathcal{P}_{i-1}$ by duplicating them $r_i$ times to generate a set of child points $\hat{\mathcal{P}}_i$, and then spread $\hat{\mathcal{P}}_i$ to the neighborhood of the parent points. 
To achieve this, we take the inspiration from \cite{yuan2018pcn} to predict the \emph{point displacement} $\Delta \mathcal{P}_i$ of $\hat{\mathcal{P}}_i$.
Then, $\hat{\mathcal{P}}_i$ is updated as $\mathcal{P}_i = \hat{\mathcal{P}}_i + \Delta \mathcal{P}_i$, where $\mathcal{P}_i$ is the output of the $i$-th SPD.

In detail, taking the shape code $\mathbf{f}$ from feature extraction, the SPD first extracts the per-point feature $\mathcal{Q}_{i-1} = \{\mathbf{q}_j^{i-1}\}_{j=1}^{N_{i-1}}$ for $\mathcal{P}_{i-1}$ by adopting the basic PointNet \cite{qi2017pointnet} framework. Then, $\mathcal{Q}_{i-1}$ is sent to the skip-transformer to learn the \emph{shape context feature}, denoted as $\mathcal{H}_{i-1} = \{\mathbf{h}_j^{i-1}\}_{j=1}^{N_{i-1}}$. 
Next, $\mathcal{H}_{i-1}$ is up-sampled by \emph{point-wise splitting} operation and duplication, respectively, where the former serves to add variations and the latter preserves shape context information. 
Finally, the up-sampled feature with size of $N_i \times 2 C'$ is fed to MLP to produce the \emph{displacement feature} $\mathcal{K}_i = \{\mathbf{k}_j^i\}_{j=1}^{N_i}$ of current step. Here, $\mathcal{K}_i$ is used for generating the point displacement $\Delta \mathcal{P}_i$, and will be fed into the next SPD. $\Delta \mathcal{P}_i$ is formulated as 
%
\begin{equation}\small
\Delta \mathcal{P}_i = \mathrm{tanh}(\mathrm{MLP}(\mathcal{K}_i)),
\label{eq:displacement}
\end{equation}
where $\mathrm{tanh}$ is the hyper-tangent activation.
\begin{figure}
\begin{center}
   \includegraphics[width=0.8\linewidth]{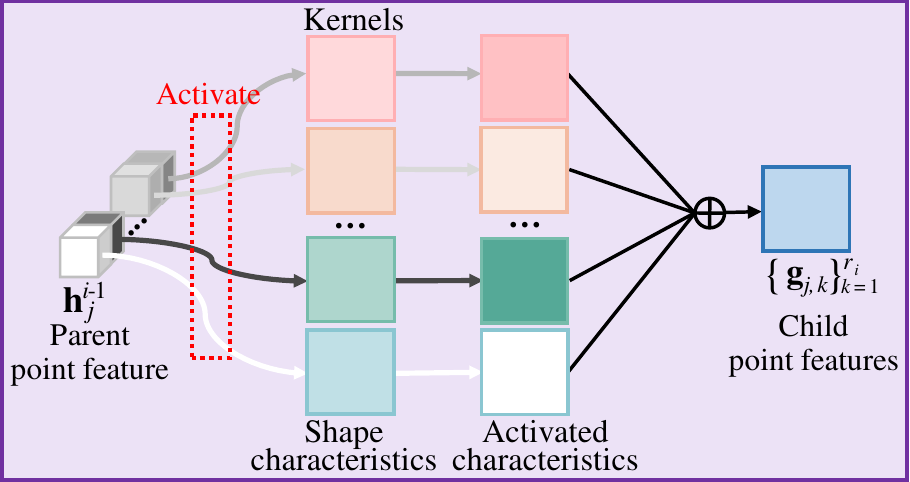}
\end{center}
    \vspace{-2mm}
   \caption{The point-wise splitting operation. The cubes are logits of the parent point feature that represent activation status of the corresponding shape characteristics (Kernels), and child point features are obtained by adding activated shape characteristics.}
\label{fig:splitting}
\vspace{-4mm}
\end{figure}

\noindent\textbf{Point-wise splitting operation.}
point-wise splitting operation aims to generate multiple child point features for each $\mathbf{h}_j^{i-1} \in \mathcal{H}_{i-1}$. 
Figure \ref{fig:splitting} shows this operation structure used in the $i$-th SPD (see Figure \ref{fig:overall}(c)). It is a special one-dimensional deconvolution strategy, where the kernel size and stride are both equal to $r_i$. 
In practice, each $\mathbf{h}_j^{i-1} \in \mathcal{H}_{i-1}$ shares the same set of kernels, and produces multiple child point features in a point-wise manner.
To be clear, we denote the $m$-th logit of $\mathbf{h}_j^{i-1}$ as $h_{j, m}^{i-1}$, and its corresponding kernel is indicated by $\mathrm{K}_m$. Technically, $\mathrm{K}_m$ is a matrix with a size of $r_i \times \mathrm{C}'$, the $k$-th row of $\mathrm{K}_m$ is denoted as $\mathbf{k}_{m, k}$, and the $k$-th child point feature $\mathbf{g}_{j, k}$ is given by
\begin{equation}\small
\vspace{-2mm}
\mathbf{g}_{j, k} = \sum_m h_{j, m}^{i-1} \mathbf{k}_{m, k}
\label{eq:dilation}.
\end{equation}
%
%
In addition, in Figure \ref{fig:splitting}, we assume that each learnable kernel $\mathrm{K}_m$ indicates a certain shape characteristic, which describes the geometry and structure of 3D shape in local region. Correspondingly, every logit $h_{j, m}^{i-1}$ indicates the activation status of the $m$-th shape characteristic. The child point features can be generated by adding the activated shape characteristics. Moreover, the point-wise splitting operation is flexible for up-sampling points. For example, when $r_i = 1$, it enables the SPD to move the point from previous step to a better position; when $r_i > 1$, it serves to expand the number of points by a factor of $r_i$.

\begin{figure}
\begin{center}
   \vspace{-3mm}
   \includegraphics[width=2.5in]{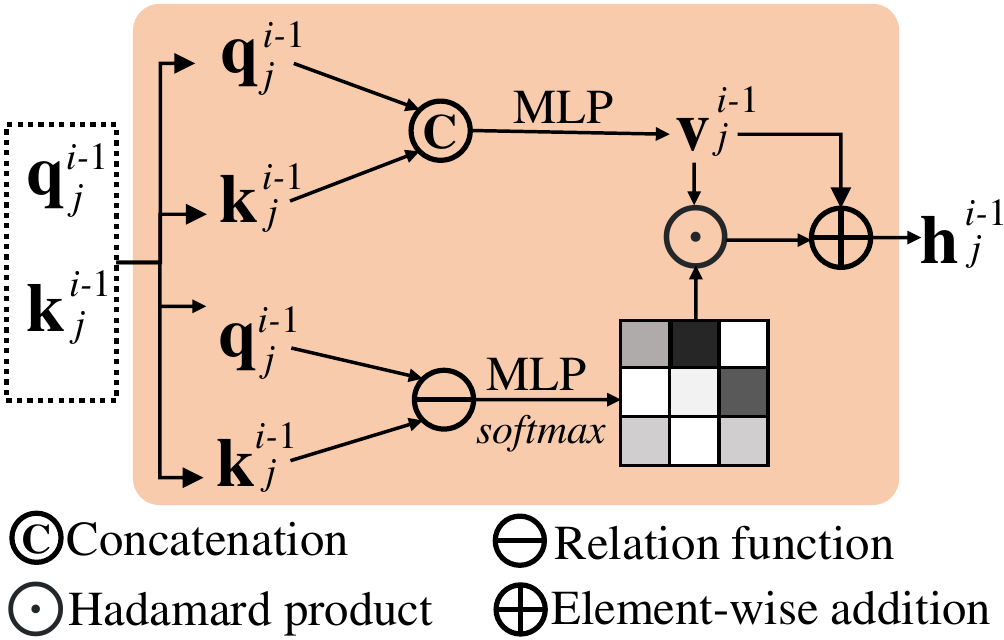}
\end{center}
   \caption{The detailed structure of skip-transformer.}
\label{fig:skip_transformer}
\vspace{-5mm}
\end{figure}

\noindent\textbf{Collaboration between SPDs. }
In Figure \ref{fig:overall}(a), we adopt three SPDs to generate the complete point cloud. We first set the up-sampling factor $r_1=1$ to explicitly rearrange seed point positions. Then, we set $r_2>1$ and $r_3 > 1$ to generate a structured tree for every point in $\mathcal{P}_1$. Collaboration between SPDs is crucial for growing the tree in a coherent manner, because information from the previous splitting can be used to guide the current one. Besides, the growth of the rooted trees should also capture the pattern of local patches to avoid overlapping with each other.
To achieve this purpose, we propose a novel \emph{skip-transformer} to serve as the cooperation unit between SPDs. In Figure \ref{fig:skip_transformer}, the skip-transformer takes per-point feature $\mathbf{q}_j^{i-1}$ as input, and combines it with displacement feature $\mathbf{k}_j^{i-1}$ from previous step to produce the shape context feature $\mathbf{h}_j^{i-1}$, which is given by
\begin{equation}\small
\mathbf{h}_j^{i-1} = \mathrm{ST}(\mathbf{k}_j^{i-1}, \mathbf{q}_j^{i-1}),
\label{eq:skip_transformer}
\end{equation}
%
where $\mathrm{ST}$ denotes the skip-transformer. The detailed structure is described as follows.

\subsection{Skip-Transformer}
Figure \ref{fig:skip_transformer} shows the structure of skip-transformer. The skip-transformer is introduced to learn and refine the spatial context between parent points and their child points, where the term ``skip'' represents the connection between the displacement feature from the previous layer and the point feature of the current layer.

Given per-point feature $\mathbf{q}_j^{i-1}$ and displacement feature $\mathbf{k}_j^{i-1}$, the skip-transformer first concatenates them. Then, the concatenated feature is fed to MLP, which generates the vector $\mathbf{v}_j^{i-1}$. Here, $\mathbf{v}_j^{i-1}$ serves as the value vector which incorporates previous point splitting information.
In order to further aggregate local shape context into $\mathbf{v}_j^{i-1}$, the skip-transformer uses $\mathbf{q}_j^{i-1}$ as the query and $\mathbf{k}_j^{i-1}$ as the key to estimate attention vector $\mathbf{a}_j^{i-1}$
, where $\mathbf{a}_j^{i-1}$ denotes how much attention the current splitting should pay to the previous one. To enable the skip-transformer to concentrate on local pattern, we calculate attention vectors between each point and its $k$-nearest neighbors ($k$-NN). The $k$-NN strategy also helps to reduce computation cost. Specifically, given the $j$-th point feature $\mathbf{q}_j^{i-1}$, the attention vector $\mathbf{a}_{j, l}^{i-1}$ between $\mathbf{q}_j^{i-1}$ and displacement features of the $k$-nearest neighbors $\{\mathbf{k}_{j, l}^{i-1} | l=1,2,\dots,k\}$ can be calculated as
\begin{equation}\small
\mathbf{a}_{j, l}^{i-1} = \frac
{\mathrm{exp}( \mathrm{MLP} ((\mathbf{q}_j^{i-1}) \ominus (\mathbf{k}_{j, l}^{i-1})  )    )}
{\sum_{l=1}^k \mathrm{exp}( \mathrm{MLP} ((\mathbf{q}_j^{i-1}) \ominus (\mathbf{k}_{j, l}^{i-1})  )    )}
\label{eq:attention_vector},
\end{equation}
where $\ominus$ serves as the relation operation, i.e. element-wise subtraction. Finally, the shape context feature $\mathbf{h}_j^{i-1}$ can be obtained by
\begin{equation}\small
\mathbf{h}_j^{i-1} = \mathbf{v}_{j}^{i-1} \oplus \sum_{l=1}^k \mathbf{a}_{j, l}^{i-1} \odot \mathbf{v}_{j, l}^{i-1}
\label{eq:h},
\end{equation}
where $\oplus$ denotes element-wise addition and $\odot$ is Hadamard product. Note that there is no previous displacement feature for the first SPD, of which the skip-transformer takes $\mathbf{q}_j^0$ as both query and key.

\begin{table*}[!t]\small
\centering
\caption{Point cloud completion on PCN dataset in terms of per-point L1 Chamfer distance $\times 10^{3}$ (lower is better).}
\begin{tabular}{l|c|cccccccc}
\toprule
Methods &Average  &Plane    &Cabinet  &Car   &Chair   &Lamp   &Couch    &Table    &Boat   \\ 
\midrule
FoldingNet  \cite{yang2018foldingnet}   &14.31  &9.49    &15.80    &12.61    &15.55   &16.41    &15.97    &13.65    &14.99   \\
TopNet  \cite{tchapmi2019topnet}     &12.15   &7.61   &13.31   &10.90    &13.82    &14.44      &14.78   &11.22  &11.12   \\
AtlasNet \cite{groueix2018atlasnet}   &10.85   &6.37    &11.94    &10.10    &12.06    &12.37    &12.99    &10.33    &10.61   \\
PCN  \cite{yuan2018pcn}   &9.64 &5.50    &22.70    &10.63    &8.70    &11.00    &11.34    &11.68    &8.59   \\
GRNet \cite{xie2020grnet}  &8.83   &6.45   &10.37   &9.45    &9.41    &7.96      &10.51   &8.44  &8.04   \\
CDN \cite{wang2020cascaded}  &8.51   &4.79   &9.97   &8.31    &9.49    &8.94      &10.69   &7.81  &8.05   \\
PMP-Net \cite{wen2020pmp} &8.73  &5.65    &11.24    &9.64    &9.51    &6.95    &10.83    &8.72   &7.25 \\
NSFA \cite{zhang2020detail}  &8.06 &4.76 &10.18 &8.63 &8.53 &7.03 &10.53 &7.35 &7.48 \\
\midrule
Ours	&\textbf{7.21} &\textbf{4.29}	&\textbf{9.16}	&\textbf{8.08}	&\textbf{7.89}	&\textbf{6.07}	&\textbf{9.23}	&\textbf{6.55}	&\textbf{6.40} \\
\bottomrule
\end{tabular}
\label{table:pcn}
\end{table*}

\begin{figure*}[!t]
\begin{center}
   \includegraphics[width=1.0\textwidth]{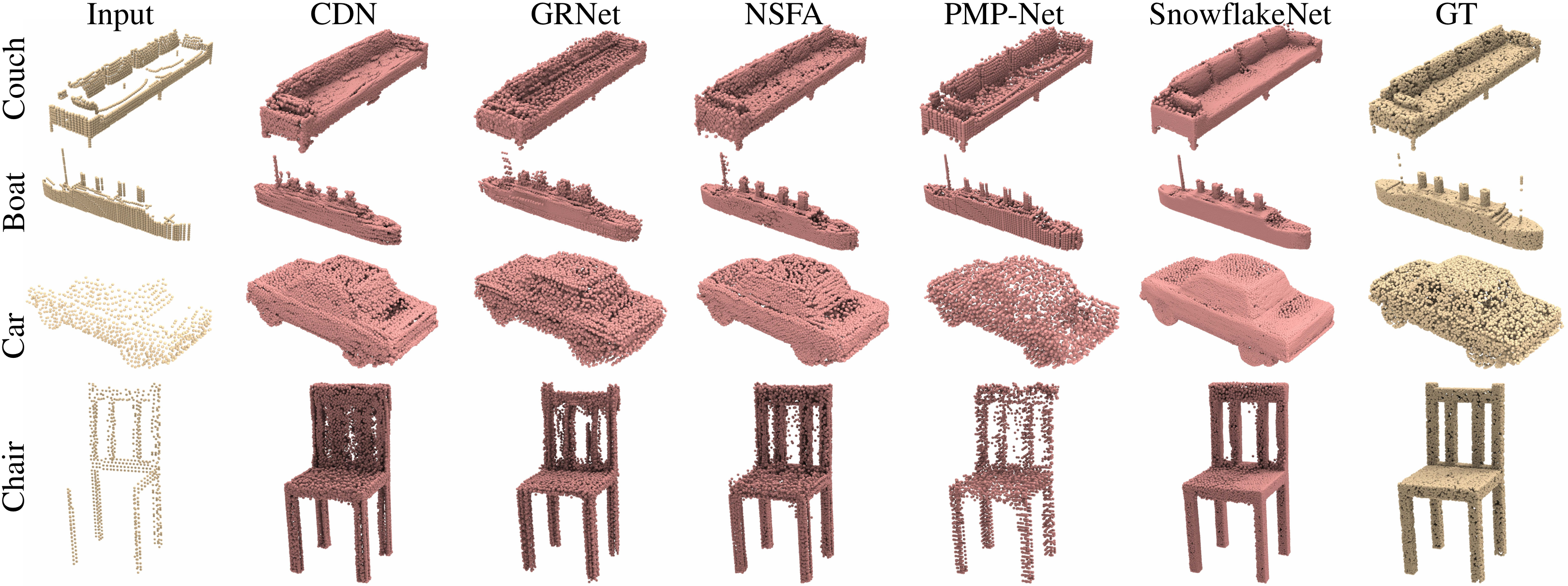}
   \caption{Visual comparison of point cloud completion on PCN dataset. Our SnowflakeNet can produce smoother surfaces (e.g. car) and more detailed structures (e.g. chair back) compared with the other state-of-the-art point cloud completion methods.}
\label{fig:pcn_vis}
\end{center}
\vspace{-6mm}
\end{figure*}

\subsection{Training Loss}

In our implementation, we use Chamfer distance (CD) as the primary loss function. To explicitly constrain point clouds generated in the seed generation and the subsequent splitting process, we down-sample the ground truth point clouds to the same sampling density as $\{\mathcal{P}_c, \mathcal{P}_1, \mathcal{P}_2, \mathcal{P}_3 \}$ (see Figure \ref{fig:overall}), where we define the sum of the four CD losses as the \emph{completion loss}, denoted by $\mathcal{L}_\mathrm{completion}$.
Besides, we also exploit the \textit{partial matching loss} from \cite{wen2020cycle} to preserve the shape structure of the input point cloud. It is an unidirectional constraint which aims to match one shape to the other without constraining the opposite direction. Because the partial matching loss only requires the output point cloud to partially match the input, we take it as the \emph{preservation loss} $\mathcal{L}_\mathrm{preservation}$, and the total training loss is formulated as
\begin{equation}\small
\mathcal{L} = \mathcal{L}_\mathrm{completion} + \lambda \mathcal{L}_\mathrm{preservation}.
\end{equation}
The arrangement is detailed in  \emph{Supplementary Material}.

\section{Experiments}
To fully prove the effectiveness of our SnowflakeNet, we conduct comprehensive experiments under two widely used benchmarks: PCN \cite{yuan2018pcn} and Completion3D \cite{tchapmi2019topnet}, both of which are subsets of the ShapeNet dataset. The experiments demonstrate that our method has superiority over the state-of-the-art point cloud completion methods. 

\subsection{Evaluation on PCN Dataset}
\noindent\textbf{Dataset briefs and evaluation metric. }
The \textit{PCN} dataset \cite{yuan2018pcn} is a subset with 8 categories derived from ShapeNet dataset \cite{chang2015shapenet}. The incomplete shapes are generated by back-projecting complete shapes into 8 different partial views. For each complete shape, 16384 points are evenly sampled from the shape surface. We follow the same split settings with PCN \cite{yuan2018pcn} to fairly compare our SnowflakeNet with other methods. For evaluation, we adopt the L1 version of Chamfer distance, which follows the same practice as previous methods \cite{yuan2018pcn}.

\noindent\textbf{Quantitative comparison.} Table \ref{table:pcn} shows the results of our SnowflakeNet and other completion methods on PCN dataset, from which we can find that SnowflakeNet achieves the best performance over all counterparts. Especially, compared with the result of the second-ranked NSFA \cite{zhang2020detail}, SnowflakeNet reduces the average CD by 0.85, which is 10.5\% lower than the NSFA's results (8.06 in terms of average CD). Moreover, SnowflakeNet also achieves the best results on all categories in terms of CD, which proves the robust generalization ability of SnowflakeNet
for completing shapes across different categories. In Table \ref{table:pcn}, both CDN \cite{wang2020cascaded} and NSFA \cite{zhang2020detail} are typical point cloud completion methods, which adopt a coarse-to-fine shape decoding strategy and model the generation of points as a hierarchical rooted tree. Compared with these two methods, our SnowflakeNet also adopts the same decoding strategy but achieves much better results on PCN dataset. Therefore, the improvements should credit to the proposed SPD layers and skip-transformer in SnowflakeNet, which helps to generate points in local regions in a locally structured pattern.

\begin{table*}[!t]\small
\centering
\caption{Point cloud completion on Completion3D in terms of per-point L2 Chamfer distance $\times 10^{4}$ (lower is better).}
\begin{tabular}{l|c|cccccccc}
\toprule
Methods &Average  &Plane    &Cabinet  &Car   &Chair   &Lamp   &Couch    &Table    &Boat   \\ 
\midrule
FoldingNet  \cite{yang2018foldingnet}   &19.07  &12.83    &23.01    &14.88    &25.69    &21.79    &21.31    &20.71    &11.51   \\
PCN  \cite{yuan2018pcn}   &18.22 &9.79    &22.70    &12.43    &25.14    &22.72    &20.26    &20.27    &11.73   \\
PointSetVoting \cite{pointsetvoting}  &18.18 &6.88    &21.18    &15.78    &22.54    &18.78    &28.39    &19.96    &11.16   \\
AtlasNet \cite{groueix2018atlasnet}   &17.77   &10.36    &23.40    &13.40    &24.16    &20.24    &20.82    &17.52    &11.62   \\
SoftPoolNet \cite{wang2020softpoolnet} &16.15   &5.81   &24.53   &11.35    &23.63    &18.54      &20.34   &16.89  &7.14   \\
TopNet  \cite{tchapmi2019topnet}     &14.25   &7.32   &18.77   &12.88    &19.82    &14.60      &16.29   &14.89  &8.82   \\
SA-Net \cite{wen2020sa}  &11.22   &5.27   &14.45   &7.78    &13.67    &13.53      &14.22   &11.75  &8.84   \\
GRNet \cite{xie2020grnet}  &10.64   &6.13   &16.90   &8.27    &12.23    &10.22      &14.93   &10.08  &5.86   \\
PMP-Net \cite{wen2020pmp} &9.23  &3.99    &14.70    &8.55    &10.21    &9.27 &12.43    &8.51   &5.77 \\ 
\midrule
Ours &\textbf{7.60} &\textbf{3.48}	&\textbf{11.09}	&\textbf{6.9}	&\textbf{8.75}	&\textbf{8.42}	&\textbf{10.15}	&\textbf{6.46}	&\textbf{5.32}	\\
\bottomrule
\end{tabular}
\label{table:completion3d}
\vspace{-4mm}
\end{table*}

\noindent\textbf{Visual comparison.} We typically choose top four point cloud completion methods from Table \ref{table:pcn}, and visually compare SnowflakeNet with these methods in Figure \ref{fig:pcn_vis}. The visual results show that SnowflakeNet can predict the complete point clouds with much better shape quality. For example, in the car category, the point distribution on the car's boundary generated by SnowflakeNet is smoother and more uniform than other methods. As for the chair category, SnowflakeNet can predict more detailed and clear structure of the chair back compared with the other methods, where CDN \cite{wang2020cascaded} almost fails to preserve the basic structure of the chair back, while the other methods generate lots of noise between the columns of the chair back.

\subsection{Evaluation on Completion3D Dataset}\label{sec:compleiton3d}
\noindent\textbf{Dataset briefs and evaluation metric.}
The Completion3D dataset contains 30958 models from 8 categories, of which both partial and ground truth point clouds have 2048 points. We follow the same train/validation/test split of Completion3D to have a fair comparison with the other methods, where the training set contains 28974 models, validation and testing set contain 800 and 1184 models, respectively. For evaluation, we adopt the L2 version of Chamfer distance on testing set to align with previous studies.

\noindent\textbf{Quantitative comparison. } In Table \ref{table:completion3d}, we show the quantitative results of our SnowflakeNet and the other methods on Completion3D dataset. All results are cited from the online public leaderboard of Completion3D\footnote{https://completion3d.stanford.edu/results}. From Table \ref{table:completion3d}, we can find that our SnowflakeNet achieves the best results over all methods listed on the leaderboard. Especially, compared with the state-of-the-art method PMP-Net \cite{wen2020pmp}, SnowflakeNet significatly reduces the average CD by 1.63, which is 17.3\% lower than the PMP-Net (9.23 in terms of average CD). On the Completion3D dataset, SnowflakeNet outperforms the other methods in all categories in terms of per-category CD. Especially in the cabinet category, SnowflakeNet reduces the per-category CD by 3.61 compared with the second-ranked result of PMP-Net. Compared with PCN dataset, the point cloud in Completion3D dataset is much sparser and easier to generate. Therefore, a coarse-to-fine decoding strategy may have less advantages over the other methods. Despite of this, our SnowflakeNet still achieves superior performance over folding-based methods including SA-Net \cite{wen2020sa} and FoldingNet \cite{yang2018foldingnet}, and we are also the best among the coarse-to-fine methods including TopNet \cite{tchapmi2019topnet} and GRNet \cite{xie2020grnet}. In all, the results on Completion3D dataset demonstrate the capability of SnowflakeNet for predicting high-quality complete shape on sparse point clouds.

\begin{figure}[h]
\begin{center}
   
   \includegraphics[width=\linewidth]{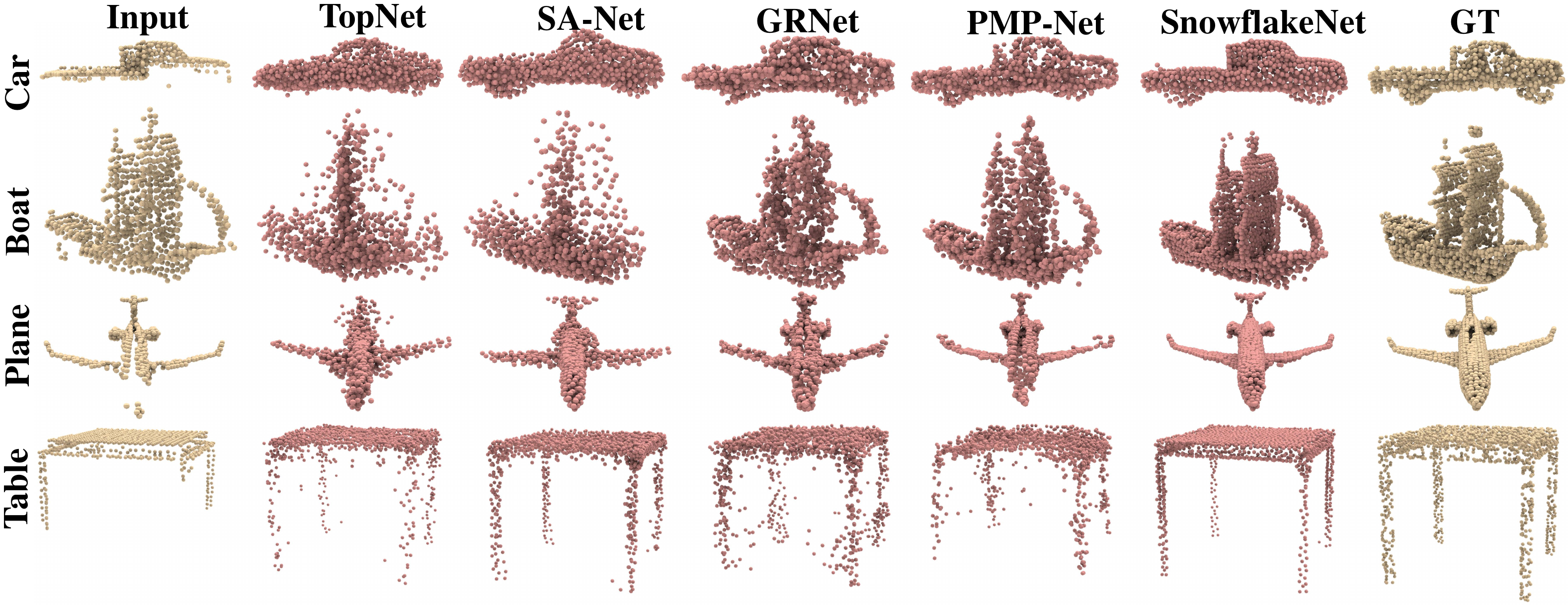}
\end{center}
   \vspace{-1mm}
   \caption{Visual comparison of point cloud completion on Completion3D dataset. Our SnowflakeNet can produce smoother surfaces (e.g. car and table) and more detailed structures compared with the other state-of-the-art point cloud completion methods.}
\label{fig:completion3d_vis}
\vspace{-5mm}
\end{figure}

\noindent\textbf{Visual comparison.} Same as the practice in PCN dataset, we also visually compare SnowflakeNet with the top four methods in Table \ref{table:completion3d}. 
Visual comparison in Figure \ref{fig:completion3d_vis} demonstrates that our SnowflakeNet also achieves much better visual results than the other counterparts on sparse point cloud completion task. Especially, in plane category, SnowflakeNet predicts the complete plane which is almost the same as the ground truth, while the other methods fail to reveal the complete plane in detail. 
The same conclusion can also be drawn from the observation of car category. In the table and boat categories, SnowflakeNet produces more detailed structures compared with the other methods, e.g. the sails of the boat and the legs of the table.

\subsection{Ablation studies}
We analyze the effectiveness of each part of SnowflakeNet. For convenience, we conduct all experiments on the validation set of Completion3D dataset. By default, all the experiment settings and the network structure remains the same as Section \ref{sec:compleiton3d}, except for the analyzed part.

\noindent\textbf{Effect of skip-transformer.} To evaluate the effectiveness of skip-transformer used in SnowflakeNet, we develop three network variations as follows. (1) The \emph{Self-att} variation replaces the transformer mechanism in skip-transformer with the self-attention mechanism, where the input is the point features of current layer. (2) The \emph{No-att} variation removes the transformer mechanism from skip-transformer, where the features from the previous layer of  SPD is directly added to the feature of current SPD layer. (3) The \emph{No-connect} variation removes the whole skip-transformer from the SPD layers, and thus, no feature connection is established between the SPD layers. The experiment results are shown in Table \ref{table:st_analysis}. In addition, we denote the original version of SnowflakeNet as \emph{Full} for clear comparison with the performance of each network variations. From Table \ref{table:st_analysis}, we can find that the transformer-based Full model achieves the best performance among all compare network variations. The comparison between the No-connect model and the Full model justifies the advantage of using skip-transformer between SPD layers, and the comparison between No-att model and Full model further proves the effectiveness of using transformer mechanism to learn shape context in local regions. Moreover, the comparison between Self-att model and No-att model shows that the attention based mechanism can also contribute to the completion performance.
\begin{table}[h]\small
\centering
\caption{Effect of skip-transformer.}
\begin{tabular}{l|c|cccc}
\toprule
Methods &avg.  &Couch    &Chair  &Car   &Lamp    \\ 
\midrule
Self-att &8.89	&6.04 &10.9	&9.42	&9.12	\\
No-att 	&9.30	&6.15 &11.2	&10.4	&9.38	\\
No-connect &9.39	&6.17 &11.3	&10.5	&9.51	\\
Full 	&\textbf{8.48} &\textbf{5.89} &\textbf{10.6}	&\textbf{9.32}	&\textbf{8.12}\\
\bottomrule
\end{tabular}
\label{table:st_analysis}
\vspace{-2mm}
\end{table}

\noindent\textbf{Effect of each part in SnowflakeNet.} To evaluate the effectiveness of each part in SnowflakeNet, we design four different network variations as follows. (1) The \emph{Folding-expansion} variation replaces the point-wise splitting operation with the folding-based feature expansion method \cite{yang2018foldingnet}, where the features are duplicated several times and concatenated with a 2-dimensional codeword, in order to increase the number of point features. (2) The \emph{$\rm E_{\rm PCN}$+SPD} variation 
employs the PCN encoder and our SnowflakeNet decoder. (3) The \emph{w/o partial matching} variation removes the partial matching loss. (4) The \emph{PCN-baseline} is the performance of original PCN method \cite{yuan2018pcn}, which is trained and evaluated under the same settings of our ablation study. In Table \ref{table:pw_expansion_analysis}, we report the results of the four network variations along with the default network denoted as \emph{Full}. By comparing $\rm E_{PCN}$+SPD with PCN-baseline, we can find that our SPD with skip-transformer based decoder can be potentially applied to other simple encoders, and achieves significant improvement. By comparing Folding-expansion with Full model, the better performance of Full model proves the advantage of point-wise splitting operation over the folding-based feature expansion methods. By comparing w/o partial matching with Full model, we can find that partial matching loss can slightly improve the average performance of SnowflakeNet.

\begin{figure}

\begin{center}
   \includegraphics[width=\linewidth]{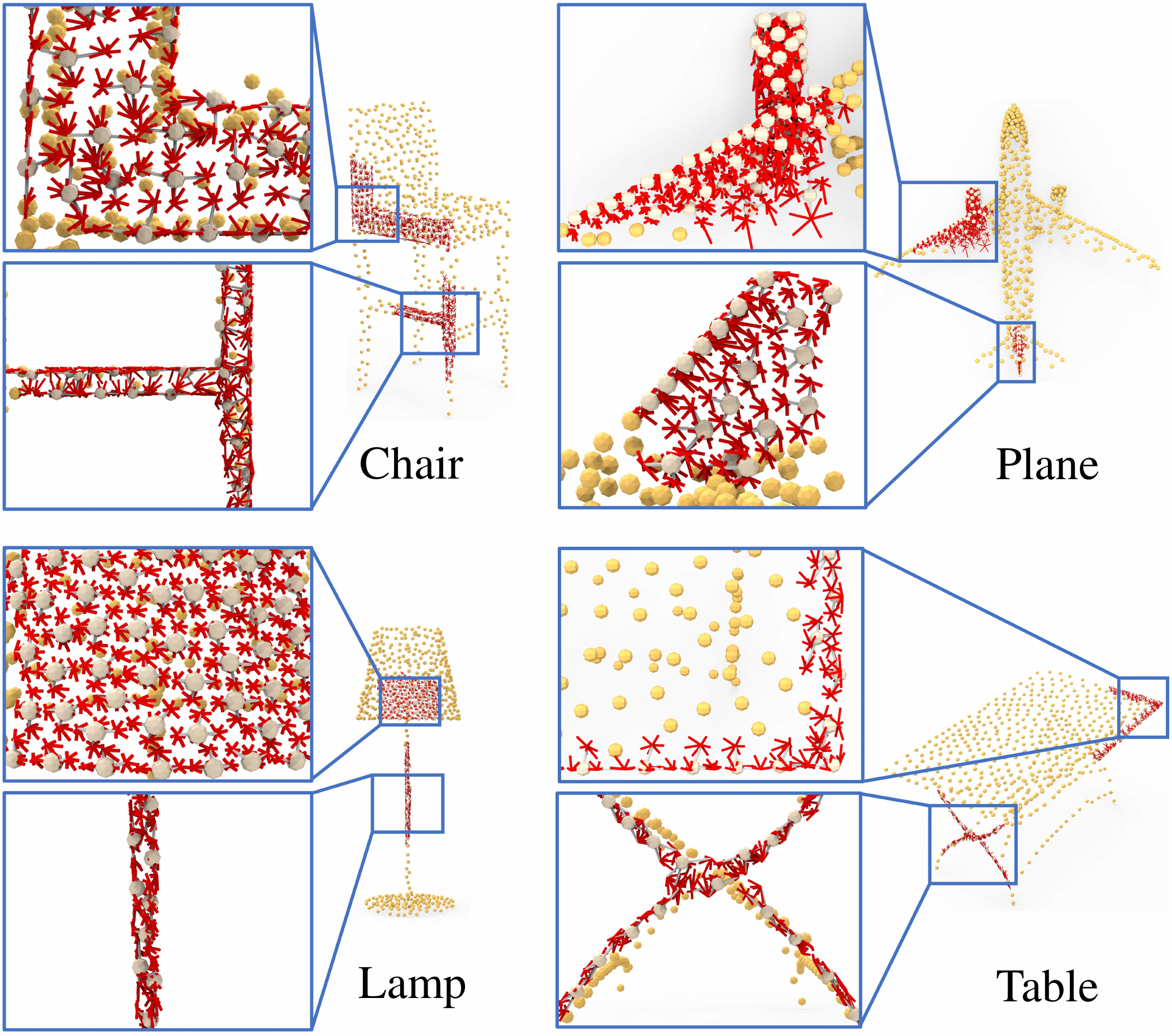}
\end{center}
    \vspace{-1mm}
    \caption{Visualization of snowflake point deconvolution on different objects. For each object, we sample two patches of points and visualize two layers of point splitting together for each sampled point. The gray lines indicate the paths of the point splitting from $\mathcal{P}_1$ to $\mathcal{P}_2$, and the red lines are splitting paths from $\mathcal{P}_2$ to $\mathcal{P}_3$.}
\label{fig:snowflake}
\vspace{-5mm}
\end{figure}

\noindent\textbf{Visualization of point generation process of SPD.} In Figure \ref{fig:snowflake}, we visualize the point cloud generation proccess of SPD. We can find that the layers of SPD generate points in a snowflake-like pattern. When generating the smooth plane (e.g. chair and lamp in Figure \ref{fig:snowflake}), we can clearly see the child points are generated around the parent points, and smoothly placed along the plane surface. On the other hand, when generating 
thin tubes and sharp edges, the child points can precisely capture the geometries.

\begin{table}\small
\centering
\caption{Effect of each part in SnowflakeNet.}
\resizebox{\linewidth}{!}{\begin{tabular}{l|c|cccc}
\toprule 
Methods &avg.  &Couch    &Chair  &Car   &Lamp \\ 
\midrule
Folding-expansion &8.80 &8.40 &10.80 &5.83 &10.10\\
$\rm E_{\rm PCN}$+SPD	&8.93 	&9.06 &11.30  &6.14	&9.23	\\
w/o partial matching &8.50 &8.72 &\textbf{10.6} &\textbf{5.78} &\textbf{8.9}\\
PCN-baseline &13.30 	&11.50   &17.00 &6.55	&18.20		\\
Full 	&\textbf{8.48} &\textbf{8.12} &10.6	&5.89	&9.32	\\
\bottomrule
\end{tabular}}
\label{table:pw_expansion_analysis}
\end{table}

\vspace{-1mm}
\section{Conclusions}
In this paper, we propose a novel neural network for point cloud completion, named SnowflakeNet. The SnowflakeNet models the generation of completion point clouds as the snowflake-like growth of points in 3D space using multiple layers of Snowflake Point Deconvolution. By further introducing skip-transformer in Snowflake Point Deconvolution, SnowflakeNet learns to generate locally compact and structured point cloud with highly detailed geometries. We conduct comprehensive experiments on sparse (Completion3D) and dense (PCN) point cloud completion datasets, which shows the superiority of our SnowflakeNet over the current SOTA point cloud completion methods.

{\small
\bibliographystyle{ieee_fullname}
\bibliography{ref}
}

\end{document}